\documentclass[letterpaper]{article} 
\usepackage{aaai2026}  
\usepackage{times}  
\usepackage{helvet}  
\usepackage{courier}  
\usepackage[hyphens]{url}  
\usepackage{graphicx} 
\usepackage{natbib}  
\usepackage{caption} 
\usepackage{algorithm}
\usepackage{algpseudocode}

\usepackage{newfloat}
\usepackage{listings}
\DeclareCaptionStyle{ruled}{labelfont=normalfont,labelsep=colon,strut=off} 
\floatstyle{ruled}
\newfloat{listing}{tb}{lst}{}
\floatname{listing}{Listing}
\usepackage{amsthm,amssymb}
\usepackage{subcaption}
\usepackage{mathtools}
\usepackage{comment}
\usepackage{booktabs, multirow} 
\usepackage{soul}
\usepackage{xcolor,colortbl} 
\usepackage{changepage,threeparttable} 
\usepackage{enumitem}
\definecolor{iccvblue}{rgb}{0.21,0.49,0.74}

\usepackage[english]{babel}

\title{Active Learning for Animal Re-Identification with Ambiguity-Aware Sampling}
\author {
    Depanshu Sani,
    Mehar Khurana,
    Saket Anand \\
}
\affiliations {
    Indraprastha Institute of Information Technology, Delhi, India \\
    \{depanshus, mehar21541, anands\}@iiitd.ac.in \\
    {\color{iccvblue}{\url{https://sites.google.com/iiitd.ac.in/aas/}}}
}   
\begin{document}

\maketitle

\begin{abstract}
Animal re-identification (Re-ID) has recently gained substantial attention in the AI research community due to its high impact on biodiversity monitoring and unique research challenges arising from environmental factors. The subtle distinguishing patterns like stripes or spots, handling new species and the inherent open-set nature make the problem even harder. To address these complexities, foundation models trained on labeled, large-scale and multi-species animal Re-ID datasets have recently been introduced to enable zero-shot Re-ID. However, our benchmarking reveals significant gaps in their zero-shot Re-ID performance for both known and unknown species. While this highlights the need for collecting labeled data in new domains, exhaustive annotation for Re-ID is laborious and requires domain expertise. Our analyses also show that existing unsupervised (USL) and active learning (AL) Re-ID methods underperform for animal Re-ID. To address these limitations, we introduce a novel AL Re-ID framework that leverages complementary clustering methods to uncover and target structurally ambiguous regions in the embedding space for mining pairs of samples that are both informative and broadly representative. Oracle feedback on these pairs, in the form of must-link and cannot-link constraints, facilitates a simple annotation interface, which naturally integrates with existing USL methods through our proposed constrained clustering refinement algorithm. Through extensive experiments, we demonstrate that, by utilizing only $0.033$\% of all possible annotations, our approach consistently outperforms existing foundational, USL and AL baselines. Specifically, we report an average improvement of $10.49$\%, $11.19$\% and $3.99$\% (mAP) on 13 wildlife datasets over foundational, USL and AL methods, respectively, while attaining state-of-the-art performance on each dataset. Furthermore, we also show an improvement of $11.09$\%, $8.2$\% and $2.06$\% (AUC ROC) for unknown individuals in an open-world setting. We also present results on 2 publicly available person Re-ID datasets, showing average gains of $7.96$\% and $2.86$\% (mAP) over existing USL and AL Re-ID methods. 
\end{abstract}

\section{Introduction}\label{sec:introduction}

\begin{figure*}
        \centering
        \includegraphics[width=0.98\linewidth]{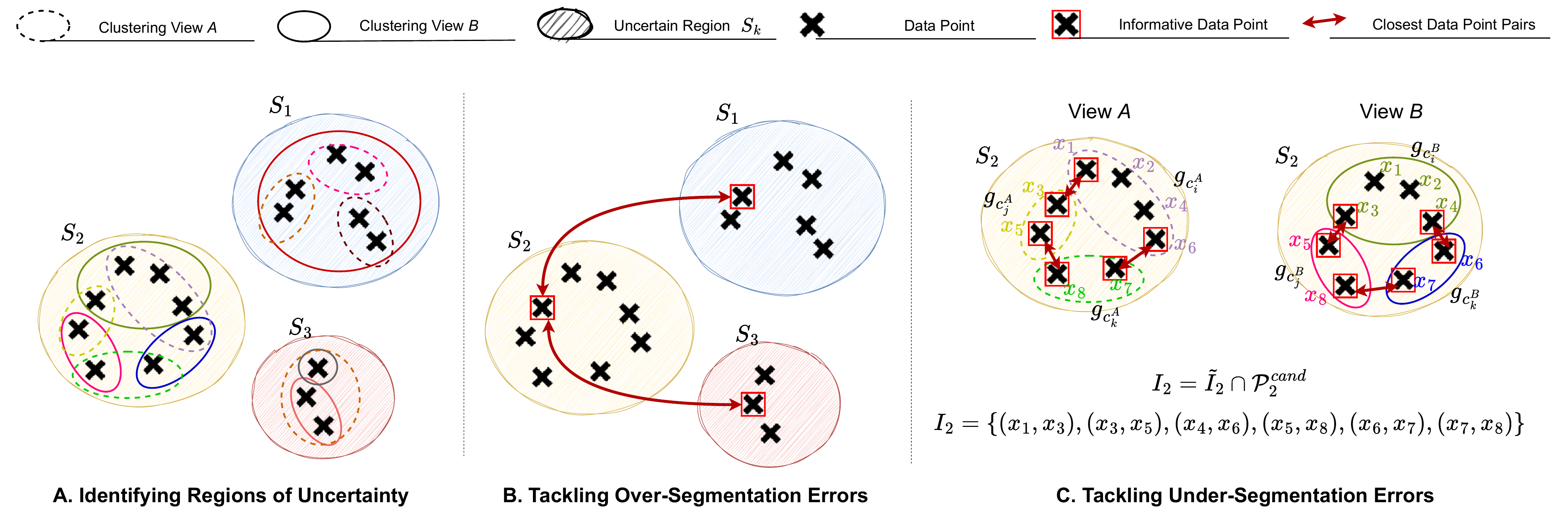}
        \caption{An illustration of Ambiguity-Aware Sampling (AAS). 
        \textbf{(A.) Identifying Regions of Uncertainty.} The disagreements between the clustering views $A$ and $B$ are used to identify regions of uncertainty, $S=\{S_1, S_2, S_3\}$, based on the transitive closure of \textit{partially overlapping} clusters from $A$ and $B$.
        \textbf{(B.) Resolving Over-Segmentation Errors.} The medoid $\mathbf{m}_k$ is chosen to be the representative of an uncertain region $S_k$. For each medoid $\mathbf{m}_k$, we obtain its nearest $k_{max}$ neighboring medoids, $\mathbf{m}_{k'} ~\forall k' \neq k$, having similarity $>s_{min}$ to form most informative image pairs.
        \textbf{(C.) Resolving Under-Segmentation Errors.} 
        $\tilde I_k$ is the set of all pairs that are in disagreement within $S_k$. Note that $\tilde{I}_k$ might contain redundant pairs that do not provide additional value if annotated. For instance, pairs $(x_1,x_6)$, $(x_2,x_6)$ and $(x_4,x_6)$ in the sub-figure represent the same ambiguity, i.e. whether the group of consistently clustered images $\{x_1, x_2, x_4\}$ should be clustered with $\{x_6\}$. 
        Hence, to filter out redundant pairs, we define $\mathcal{P}_k^{cand}$ to be a set of candidate image pairs that contains the nearest inter-cluster neighbors for both the methods, which may or may not be inconsistent.
        Therefore, for each uncertain region $S_k$, we define the set of most informative and non-redundant \textit{inconsistent} image pairs as $I_k=\tilde I_k ~\cap~ \mathcal{P}_k^{cand}$. 
        Finally, the set of informative medoid pairs ($\mathcal{U}_{os}$) and non-redundant inconsistent pairs ($\mathcal{U}_{us}$) defines our sampling pool $\mathcal{U} = \{\mathcal{U}_{os} ~\cup~ \mathcal{U}_{us}\}$ from which the image pairs are sampled for annotation.
        }
        \label{fig:sampling-overview}
\end{figure*}

Re-ID is a critical task that aims to match instances of the same individual across different images. While traditionally explored in the context of human surveillance, Re-ID has found growing relevance in wildlife monitoring, where it is used to track individual animals over time. The ability to Re-ID individuals plays a pivotal role in ecological studies, population estimation, behavioral tracking and conservation planning. Non-invasive approaches, like the use of camera trap images, often require manual processing to ensure reliable re-identification. 

Animal Re-ID has recently gained substantial attention in the AI research community, not only due to its high impact on biodiversity monitoring, but also because of its unique and significantly more complex challenges compared to other Re-ID tasks. For instance, unlike human Re-ID where the key distinguishing attributes are same for all, animal Re-ID depends on visual patterns that are \textit{species-dependent}, such as stripes for tigers, spots for cheetahs and rosettes for leopards. To address this, several AI-enabled systems have been developed for species-agnostic animal Re-ID \cite{re-id-any-animal, wildlife-datasets, multispecies-animal-reid}. At their core, these systems leverage foundation models that are trained on large-scale multi-species wildlife datasets that are manually annotated to find instances of the same individual across diverse species and geographies. 


Despite recent advances, existing foundation models for animal Re-ID \cite{wildlife-datasets, multispecies-animal-reid} struggle to perform adequately for new environments. Moreover, adapting Re-ID systems to new environments remains challenging due to the high cost of manual annotation. Manually identifying individuals in a large image set is tedious and error-prone as it needs discerning fine-grained patterns in images. A natural way to approach this is to frame supervision in terms of pairwise annotations, where annotators simply indicate whether two images are of the same individual. 
Unfortunately, this labeling process is expensive and time-consuming as it requires identifying matching pairs within a large dataset, where the number of comparisons grows quickly with the dataset size.

To mitigate the cost of exhaustive annotation, many recent unsupervised Re-ID approaches \cite{spcl, refining-pseudo-labels, cluster-contrast, camera-aware-proxies, multi-centroid-representation-network, discrepant-multi-instance-proxies, RTMem} rely on pseudo-labeling via clustering, where the model is iteratively refined using automatically generated labels from the structure of the feature space. 
However, the effectiveness of pseudo-labeling hinges critically on cluster purity. Errors in clustering can introduce harmful supervision. Broadly, clustering errors can be categorized into two types: \textit{under-segmentation}, where samples from different identities are erroneously grouped into the same cluster, and \textit{over-segmentation}, where samples from the same identity are incorrectly split into multiple clusters. Both scenarios are detrimental—under-segmentation leads to identity confusion and suppresses inter-class separability, while over-segmentation fragments identity representations and weakens intra-class consistency. These challenges highlight the fragility of unsupervised pseudo-labeling and underscore the importance of guiding the learning process using reliable supervisory signals.

This motivates the use of active learning (AL), which seeks to identify and label the most informative examples under a constrained annotation budget. In the Re-ID setting, where supervision is naturally framed through pairwise comparisons, not all image pairs are equally valuable. Most pairs offer little learning signal—being either obviously matched or mismatched. Instead, the greatest benefit lies in annotating ambiguous pairs, especially those near decision boundaries. Moreover, to ensure robust generalization and avoid overfitting, it is equally important that selected pairs are diverse, covering a wide range of identities and appearance variations.
Traditional works explored AL in the context of Re-ID using an entropy-based \cite{active-image-pair}, rank-based \cite{HITL} and reinforcement learning-based \cite{DRAL, contextual-AL} selection criterion. The primary focus of many existing solutions has been on person or vehicle Re-ID, often resulting in works that also target modeling the multi-camera setup \cite{reid-in-camera-network, consistent-aware-in-camera-network, exploiting-transitivity}. With advances in unsupervised deep clustering methods, recent methods exploit cluster statistics to identify the most informative pairs \cite{MASS} or triplets \cite{HMMN, LBAS}. While NIS \cite{nis} designs an AL strategy to estimate the population size for wildlife datasets using nested information sampling, to the best of our knowledge, there exists no prior work that leverages AL for animal Re-ID.

In this work, we argue that the pseudo-labels obtained via different clustering algorithms are often structurally distinct, reflecting their differing inductive biases. Therefore, the variations in these pseudo labels can reveal ambiguities within the feature space. By identifying the disagreements between these complementary clustering views, we aim to uncover the most uncertain image pairs that can help reduce \textit{under-segmentation} errors. Conversely, agreement between methods does not guarantee correctness, as it can hide \textit{over-segmentation} errors. Therefore, we propose a novel active sampling strategy which can be used to sample pairs for addressing both under- and over-segmentation (Fig \ref{fig:sampling-overview}). Moreover, our AL Re-ID framework is designed in a way that can be seamlessly integrated with existing unsupervised or semi-supervised deep clustering methods for efficient training. To enable this, we introduce a \textit{post-hoc} constrained clustering algorithm that can be used to incorporate the human feedback into the pseudo labels generated via the base clustering method. Our contributions are summarized below:

\begin{enumerate}
\item We propose Ambiguity-Aware Sampling (AAS) strategy that leverages disagreements between two complementary clustering algorithms—DBSCAN (density-based) and FINCH (nearest-neighbor-based)—to identify and sample most informative, uncertain and diverse pairs of images for annotation.
\item We introduce a Non-Parametric, Plug-and-Play (NP3) algorithm, that is a \textit{post-hoc} constrained clustering method and is agnostic to the underlying clustering technique used to obtain the labels. 
\end{enumerate}

By integrating these strategies into the AL Re-ID pipeline, we demonstrate the following:

\begin{enumerate}
\item Existing foundation models, unsupervised training strategies and AL Re-ID methods show limited gains over a pre-trained ResNet-50 baseline on animal Re-ID benchmarks.
\item NP3 algorithm enables seamless integration of AL Re-ID techniques with existing unsupervised training pipelines.
\item State-of-the-art performance on 13 wildlife and 2 person Re-ID benchmark datasets.
\end{enumerate}

\section{Related Work}\label{sec:related_works}
\textbf{Animal ReID:}
Ecologists have employed various Re-ID methods for decades \cite{animal-reid-survey, Tuia2022}, including tagging \cite{telemetry, satellite-tagging, biotelemetry}, scarring \cite{photo-id, photo-identification}, and banding \cite{leg-banding}. However, these methods are labor-intensive, expensive \cite{animal-reid-survey}, and can cause pain or impact animal behavior \cite{effects-of-tagging}.

Due to the active shift towards non-invasive monitoring, camera trap images powered by computer vision techniques have emerged as an efficient and scalable alternative. Early methods relied on classical techniques like SIFT \cite{sloop, wild-id, hotspotter}, but the field has advanced significantly with deep learning \cite{tiger-reid, primatefaceid}. A substantial body of research is focused on curating large-scale benchmark datasets that facilitate the design of modern algorithms \cite{animal-face-dataset, mammal-club, wildlife-datasets, wildlife-reid-10k}. At the same time, researchers have also focused on developing deep learning models for multi-species animal Re-ID. UniReID \cite{re-id-any-animal} uses pre-trained CLIP as the backbone model and learns the visual and textual guidance generators to obtain semantic knowledge, which is then used to guide their UniReID focus on salient visual content via the text-guided attentive module. MegaDescriptor \cite{wildlife-datasets} leverages the SWIN transformer architecture \cite{liu2021Swin} and is trained on ~2.8M images spanning ~30K identities. MiewID \cite{multispecies-animal-reid} is another foundation model that utilizes a simpler convolutional backbone and is trained on ~2.3M images of ~37K individuals. 
Despite the availability of such foundation models, we observe that they often underperform on unseen data, highlighting a crucial need for fine-tuning. While WildFusion \cite{WildFusion_ECCVw2024} recently proposed fusing and calibrating global (e.g. MegaDescriptor) and local (e.g. SIFT) feature matching scores, we believe fine-tuning Re-ID models on unseen data is still crucial.

\textbf{Unsupervised ReID:}
These approaches are broadly categorized as either unsupervised domain adaptive (UDA) or fully unsupervised learning (USL). A key difference is that USL methods operate on unlabeled data without relying on a source-domain dataset. SpCL \cite{spcl} is a pioneering example for both UDA and USL that uses a self-paced contrastive learning framework with a hybrid memory. This hybrid memory is crucial for generating the necessary signals to learn effective features by jointly distinguishing between source classes, target clusters and unclustered instances. Subsequent research has built on this foundation by focusing on improving cluster proxies and memory mechanisms \cite{cluster-contrast, camera-aware-proxies, multi-centroid-representation-network, discrepant-multi-instance-proxies, RTMem}, dynamically refining pseudo-labels \cite{refining-pseudo-labels, online-pseudo-label-generation} and leveraging multi-camera information to boost performance \cite{reid-in-camera-network, consistent-aware-in-camera-network, camera-aware-proxies, pseudo-label-refinement-intra-camera}. Given its inherent generalizability, we use the SpCL framework as our base training method, acknowledging that more recent USL methods could similarly be integrated.

\textbf{AL for ReID:}
Several recent efforts have explored AL strategies, tailored to the Re-ID setting, each introducing specific heuristics to select diverse informative image pairs \cite{active-image-pair, HITL, DRAL, contextual-AL, reid-in-camera-network, consistent-aware-in-camera-network, exploiting-transitivity, HMMN}. For instance, the MASS framework \cite{MASS} focuses on identifying and purifying impure clusters by forming coarse clusters using FINCH \cite{FINCH} and then querying annotations for a central pivot of the coarse cluster and its neighbors.
While effective in moderately clean settings, such strategies can waste annotation budget on highly under-segmented clusters containing multiple identities, where refinement is unlikely to succeed.
The state-of-the-art in AL for person Re-ID, LBAS \cite{LBAS}, aims to balance positive and negative supervision within the annotation budget. They first generate clusters using DBSCAN \cite{DBSCAN}, then mine informative triplets within clusters by selecting anchor-positive-negative sets using farthest point sampling. Although this approach promotes diversity in the labeled pairs, it is highly sensitive to DBSCAN’s parameters and tends to discard the detected outliers, which often represent the most ambiguous and thus most informative samples. We note that none of the recent AL Re-ID methods have open-sourced their implementation. 
Moreover, to the best of our knowledge, there exists no AL Re-ID method that is tested on wildlife datasets. The closest and the most relevant work to ours is NIS \cite{nis}, which designs an AL technique based on nested importance sampling to estimate the size of the animal population in the wildlife datasets. Specifically, it constructs a fully connected similarity graph and samples vertices with estimated low degrees, followed by sampling their most similar neighbors. However, it treats all sampled vertices as equally uncertain and uniformly allocates a fixed number of annotations per sampled vertex, leading to inefficient labeling when the ambiguity is highly variable. 


\section{Methodology}
Let $\mathcal{D} = \{x_1, \ldots, x_n\}$ be the dataset comprising of cropped images of individuals. Assume $f:\mathcal{D} \rightarrow \mathbb{R}^d$ be a feature extractor that maps an image $x_i$ to its vector representation $\mathbf{x}_i$. We employ two clustering methods, $A$ and $B$, on all the feature vectors $\mathcal{X} = \{\mathbf{x}_1, \ldots, \mathbf{x}_n\}$ to get clusters $\mathcal{C}^A = \{c_1^A, \ldots, c_{\vert \mathcal{C}^A \vert}^A\}$ and $\mathcal{C}^B = \{c_1^B, \ldots, c_{\vert \mathcal{C}^B \vert}^B\}$, respectively. Each image $x_k \in \mathcal{D}$ is assigned a pseudo-identity $\hat{y}_k \in \mathcal{C}^A$ by method $A$ and $\tilde{y}_k \in \mathcal{C}^B$ by method $B$. We denote the set of images belonging to a specific cluster $c_i^A$ as $\mathcal{Y}_{c_i^A}$, and similarly for $c_j^B$ as $\mathcal{Y}_{c_j^B}$. 

\subsection{Framework for Ambiguity-Aware Sampling} \label{sec:aas-framework}
Our goal is to create a pool of unlabeled image pairs ($\mathcal{U}$), which is a subset of all image pairs ($\mathcal{D} \times \mathcal{D}$). This pool $\mathcal{U}$ is designed to contain only the most informative pairs, identified through an analysis of agreements and disagreements between the complementary clustering methods A and B. Due to page length restrictions, we provide a brief overview of the process for constructing $\mathcal{U}$ and defining a marginal distribution over its image pairs, which subsequently guides the sampling of pairs for human annotation. Our AL Re-ID framework is summarized in Algorithm \ref{alg:ambiguity-sampling}. A detailed explanation of each step is provided in the supplementary\footnote{\textbf{Extended Version:} https://arxiv.org/abs/2511.06658}.

\noindent \textbf{A. Identifying Regions of Uncertainty via Disagreement: }
We focus on pairs of clusters with a partial overlap ($0\!<IoU\!<1$), which indicate disagreements between methods $A$ and $B$. We define a set of connected components, $S \!=\! \{S_1, \ldots, S_M\}$, where each $S_k$ is a distinct ``region of uncertainty''. These regions are the transitive closure of partially overlapping clusters, and their images are ``tangled'' as their assignments are inconsistent across the two methods (See Fig. \ref{fig:sampling-overview}A). This uncertainty primarily signals potential under-segmentation errors, where one method's cluster should be further divided based on the other's assignments. Conversely, these errors can also be viewed as over-segmentation errors for the other clustering method.

To address both under- and over-segmentation issues, our AL approach builds a sampling pool $\mathcal{U}$ containing diverse pairs both from \textit{within} (to resolve under-segmentation) and \textit{across} (to resolve over-segmentation) the identified regions of uncertainty. A key benefit of this strategy is that if the size of $\mathcal{U}$ is smaller than the annotation budget, our method can reduce labeling effort by only querying the most informative pairs. In cases where there are no disagreements, we do not actively sample and instead rely on the base USL training.

\noindent \textbf{B. Tackling Over-Segmentation Errors:}
To address over-segmentation errors, where a single identity is split across multiple uncertain regions, we construct a set of informative pairs, $\mathcal{U}_{os}$. Our goal is to find potential must-links between images in different regions that may belong to the same individual.
Our strategy samples representative pairs based on the medoids of these regions of uncertainty. For each region $S_k$, its medoid is the image whose feature vector is closest to all other feature vectors within that region. From the collection of all medoids $\{\mathbf{m}_1, \ldots, \mathbf{m}_M\}$, we generate candidate pairs for $\mathcal{U}_{os}$ by identifying the $k_{\max}$ nearest neighbors for each medoid in the feature space, as shown in Fig. \ref{fig:sampling-overview}(B). A pair of images is included in $\mathcal{U}_{os}$ only if their similarity is greater than or equal to a predefined threshold, $s_{\min}$. This threshold is crucial for filtering out obvious mismatches and focusing on pairs that are potentially from the same identity.

\noindent \textbf{C. Tackling Under-Segmentation Errors: }
To identify informative pairs for resolving under-segmentation errors, we focus on fine-grained disagreements between clustering methods $A$ and $B$. Our goal is to find image pairs that can resolve whether they should be assigned to different clusters.

First, we define $\tilde{I}_k$ as the set of all inconsistent pairs in a region $S_k$, using the symmetric difference ($\triangle$) between the pairs grouped by method $A$ ($A_k^+$) and method $B$ ($B_k^+$):
\begin{equation} \label{eq:inconsistent-pairs}
    \tilde{I}_k = A_k^+ ~\triangle~ B_k^+
\end{equation}
This set, however, can contain redundant pairs that resolve the same ambiguity. To create a more efficient set of queries, we propose resolving each ambiguity by querying only the single, most informative pair. The most ambiguous boundaries are represented by the ``closest pairs'' between distinct clusters. We define $\mathcal{P}_k^{\text{cand}}$ as the set of these closest pairs for all distinct cluster pairs within $S_k$ (for both A and B):
\begin{align}
    \nonumber\mathcal{P}_k^{\text{cand}} = \bigg\{&\underset{(x_u, x_v)}{\arg \min} \left(\text{dist}(x_u, x_v)\right) \mid (x_u, x_v) \in \mathcal{Y}_{c_i^\zeta} \times \mathcal{Y}_{c_j^\zeta} \\ 
     & \forall c_i^\zeta, c_j^\zeta \in S_k, i \neq j \text{ and } \zeta \in \{A, B\}\bigg\}
\end{align}
We then filter the set of initial inconsistent pairs, $\tilde{I}_k$, by intersecting it with our set of closest candidate pairs (Fig. \ref{fig:sampling-overview}(C)):
\begin{equation} \label{eq:filtered-inconsistent-pairs}
    I_k = \tilde{I}_k ~\cap~ \mathcal{P}_k^{\text{cand}} \quad \forall~ k \in \{1, \ldots, M\}
\end{equation}
Finally, the sampling pool for under-segmentation, $\mathcal{U}_{us}$, is the union of all $I_k$ across all uncertainty regions, comprising the most informative and non-redundant pairs.

\noindent \textbf{D. Defining Marginal Probability Distribution over $\mathcal{U}$:}
Our objective is to define the marginal probability distribution $P(Y)$ over any pair of images $Y=(x_u, x_v)$ in the total sampling pool $\mathcal{U}\! =\! \mathcal{U}_{os} \cup\; \mathcal{U}_{us}$. This distribution is a weighted combination of our strategies for resolving over- and under-segmentation errors, with a prior probability $\epsilon \in [0, 1]$ controlling their relative importance.

For pairs from the over-segmentation pool $\mathcal{U}_{os}$, the conditional probability is their feature similarity:
\begin{equation}
P(\!Y\!=\!(x_u, x_v)\! \mid\! Y\! \in\! \mathcal{U}_{os}\!)\! =\! \frac{\text{sim}(x_u, x_v)}{\sum \text{sim}(x_u', x_v')}
\end{equation}
where the sum in the denominator runs $\forall {(x_u',x_v') \in \mathcal{U}_{os}}$. For pairs from the under-segmentation pool $\mathcal{U}_{us}$, we recognize that the relevance of an inconsistent pair depends on its relationship—whether it's an \textit{inlier--inlier} $(\beta_{1})$, \textit{inlier--outlier} $(\beta_{2})$ or \textit{outlier--outlier} $(\beta_{3})$ pair, as identified by DBSCAN. We model the conditional probability for these pairs as a product of three factors: a prior probability ($\pi_i$) for the relationship type, a likelihood ($\rho_{k|i}$) of sampling from the uncertain region and a conditional probability ($\omega_{y|ik}$) of sampling the specific pair itself. This approach prioritizes regions with more ambiguous underlying structures. These probabilistic components are formally defined in the supplementary. The conditional probability of sampling a pair from $\mathcal{U}_{us}$ is:
\begin{equation}
\label{eq:prob_us_final}
P(Y=(x_u, x_v) \mid Y \in \mathcal{U}_{us}) = \pi_i \cdot \rho_{k|i} \cdot \omega_{y|ik}
\end{equation}
The final marginal probability distribution for $Y\! \in\! \mathcal{U}$ is:
\begin{equation}
\label{eq:marginal}
P(Y) \!= \epsilon\! \cdot\! P(Y\! \mid\! Y\! \in\! \mathcal{U}_{os}) + (1\! -\!\epsilon)\! \cdot\! P(Y\! \mid\! Y\! \in\! \mathcal{U}_{us}) 
\end{equation}

\begin{figure*}
        \centering
        \includegraphics[width=0.92\linewidth]{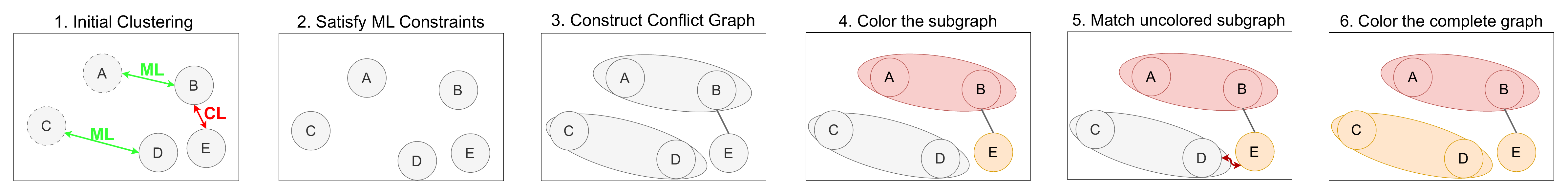}
        \caption{Illustration of the proposed Non-Parametric, Plug-and-Play (NP3) algorithm for refining pseudo labels using pairwise constraints. Given an arbitrary initial clustering, NP3's first step is to satisfy all ML constraints by merging clusters that contain ML pairs. While this addresses ML constraints, it might lead to violations of CL constraints. To purify such impure clusters, our algorithm leverages a combination of graph coloring and Hungarian matching. Intuitively, all ML pairs inside an impure cluster are treated as single conceptual nodes for graph coloring, and any remaining CL pairs in the impure cluster are considered connected via edges. The graph is then colored, ensuring that all CL-constrained pairs are assigned different ``colors'' and thus separated. For any data points that remain unconstrained by either ML or CL pairs, a Hungarian matching is employed to associate them with their closest existing cluster, a strategy designed to minimize the overall change in the cluster structure.}
        \label{fig:NP3-example}
\end{figure*}

\noindent\textbf{E. Refining Pseudo Labels with Pairwise Constraints:}
To address the limitations of existing constrained clustering methods (detailed in supp.), particularly the lack of general-purpose methods, we introduce a Non-Parametric, Plug-and-Play (NP3) algorithm that refines any given cluster partitions using pairwise constraints. Our method first enforces must-link ($ML$) constraints by merging clusters, followed by a systematic purification process that is used to resolve any resulting cannot-link ($CL$) violations.

For each impure cluster, we first consider only the data points and constraints within that cluster. Using only these internal constraints, we define a $ML$ group as a set of points that must be clustered together, based on the transitive closure of the $ML$ constraints. We then build a conflict graph where each node represents one of these local $ML$ groups and an edge connects any two groups with a $CL$ constraint between them. The connected components of this conflict graph are our $CL$ groups for the impure cluster.

Since a conflict graph's chromatic number represents the minimum number of partitions needed to satisfy all $ML$ and $CL$ constraints, we can use graph coloring to purify our clusters. We therefore assign cluster labels to the cannot-link group with the highest chromatic number by coloring, and use a Hungarian matching algorithm to assign labels to the remaining cannot-link groups of the impure cluster. This process, which is detailed in the supplementary, results in a final clustering that satisfies all constraints (Fig. \ref{fig:NP3-example}).

\noindent \textbf{Key Takeaways:}
Our core idea is to sample \textbf{diverse} pairs to address both \textbf{under-} and \textbf{over-segmentation} errors. We achieve this by first identifying ``uncertain regions'' ($S$) that span the entire feature space. Our sampling pool, $\mathcal{U}$, is composed of pairs both \textit{within} ($\mathcal{U}_{us}$) and \textit{across} ($\mathcal{U}_{os}$) these regions, which promotes diversity across the feature space.
\begin{itemize}
    \item \textbf{Over-segmentation} is addressed by querying pairs from $\mathcal{U}_{os}$ to determine if nearby images labeled as different are actually the same.
    \item \textbf{Under-segmentation} is addressed by querying pairs from $\mathcal{U}_{us}$ to determine if neighboring images merged by one method should, in fact, be separate.
\end{itemize}
This dual-sampling strategy covers most clustering ambiguities while maintaining diverse data coverage.

\begin{algorithm}[h]
\caption{AAS Framework for Training Re-ID Models}\label{alg:ambiguity-sampling}
\begin{algorithmic}[1]
\Require Unlabeled images $\mathcal{D}$, feature extractor $f$, clustering algorithms $A, B$, budget $\mathcal{B}$
\State Extract features $\mathcal{X} = \{f(x_i) \mid x_i \in \mathcal{D}\}$.
\State \textbf{Step 1: Ambiguity-Aware Sampling}
    \begin{enumerate}[label=(\alph*)]
        \item Generate clusters $\mathcal{C}^A=A(\mathcal{X})$ and $\mathcal{C}^B=B(\mathcal{X})$, and apply existing ML/CL constraints.
        \item Identify `regions of uncertainty' $S=\{S_1, \ldots, S_M\}$ from disagreements between $\mathcal{C}^A$ and $\mathcal{C}^B$.
        \item Construct sampling pools for over-segmentation ($\mathcal{U}_{os}$) and under-segmentation ($\mathcal{U}_{us}$).
        \item Define a marginal probability distribution over the total pool $\mathcal{U} = \mathcal{U}_{os} \cup \mathcal{U}_{us}$ (Eq. \ref{eq:marginal}).
        \item Sample $\mathcal{B}$ pairs for annotation using this distribution.
    \end{enumerate}
\State \textbf{Step 2: Constrained Pseudo-Label Refinement}
    \begin{enumerate}[label=(\alph*)]
        \item Generate pseudo-labels using a base USL method.
        \item Enforce ML constraints by merging clusters.
        \item For each resulting impure cluster, construct a conflict graph of must-link groups.
        \item Resolve CL violations by graph coloring and Hungarian matching 
    \end{enumerate}
\State \textbf{Step 3: Model training} Retrain the Re-ID model with refined pseudo-labels using base USL method.
\end{algorithmic}
\end{algorithm}

\section{Experimental Setup} \label{sec:experiments}
\textbf{Datasets:}
We tested our method on a collection of 13 wildlife datasets, all sourced from a subset of WildlifeReID-10k \cite{wildlife-reid-10k}, while ensuring that none of these datasets were used to train the foundation models. 
Dataset details are deferred to the supplementary, including evaluations on two human Re-ID tasks using the Market-1501 and Person-X datasets for completeness.

\textbf{Experimental Protocol:}
Prior animal Re-ID studies \cite{wildlife-datasets, wildlife-reid-10k} used the full training set as the gallery and the test set as the query. This protocol enabled \textit{open-set} scenarios, where the query could contain individuals not in the gallery. In \cite{multispecies-animal-reid}, only the test set was used for generating both the gallery and the query set. 
Since the open-set setting is more realistic for an Animal Re-ID task, we maintain the test set as query and create the gallery set by choosing exemplary images from a subset of the training set. This strategy enables an open-set evaluation as unseen individuals in the test set are absent in the gallery. We create a held out set of randomly chosen 20\% of the individuals in the training set, and up to five exemplars per ID were selected from the remaining 80\% of the individuals, using MegaDescriptor based similarity. For a fair comparison, we follow the same protocol for every method. Further details are provided in the supplementary.                                                                       


\textbf{Evaluation Metrics:}
For evaluating animal Re-ID performance in a closed-set scenario, we employed conventional metrics typically used in person Re-ID: mean Average Precision (mAP) and Top-$\{1, 3, 5, 10\}$ accuracy. Additionally, we included mean Inverse Negative Penalty (mINP) for Animal Re-ID. The Top-$k$ accuracy was computed independently for each dataset before being averaged.
\cite{wildlife-reid-10k} noted that the datasets for animal species might be highly imbalanced, both in terms of number of images as well as individuals. They proposed balancing the performance by averaging over all individuals and datasets to provide a general Re-ID performance measure for each individual. Following the same setting, we report Balanced Accuracy on Known Samples (BAKS) to measure the Re-ID performance in closed setting. Similarly, Balanced Accuracy on Unknown Samples (BAUS) uses a fixed threshold on Re-ID scores to determine whether the query is `known' or `unknown'. Since a single threshold may not fully reflect overall open-set performance, we instead report the AUC-ROC. 

\textbf{Baselines:}
We benchmarked the animal Re-ID datasets using two foundation animal Re-ID models (MegaDescriptor \cite{wildlife-datasets} and MiewID \cite{multispecies-animal-reid}) and multiple USL methods (ICE \cite{ice_iccv2021}, PPLR \cite{pplr_cvpr2022}, ClusterContrast \cite{cluster-contrast}, RTMem \cite{RTMem}, PurificationReID \cite{purificationreid_tip2023} and SpCL \cite{spcl}). As mentioned previously, we used SpCL \cite{spcl} as our base USL method for training Re-ID models. For benchmarking existing AL Re-ID methods, due to lack of open-sourced implementation, we re-implemented the most recent and current state-of-the-art AL person Re-ID method, GBAS and LBAS, proposed by \cite{LBAS}. We also used a simple baseline that actively samples random pairs of unlabeled images. We also used NIS \cite{nis}, which is not an AL Re-ID method but was proposed as an AL technique designed for animal population estimation. We used the NIS sampling strategy to get oracle feedback for pairwise ML and CL constraints. Similar to all previous USL and AL Re-ID methods, we also use a ResNet-50 backbone to extract the features. We also report baseline (pretrained) and skyline (fine-tuned on 100\% data) performances.

\textbf{Implementation Details:}
We use the official code repositories for all foundation, USL and AL Re-ID (only NIS available) methods. Since our AL Re-ID is based on SpCL, we use their official training configuration for training our Re-ID model. This implies that our similarity function $\text{sim}(x_u, x_v)$ is based on the reciprocal K-Nearest Neighbors, as in SpCL. Instead of retraining from scratch after each AL cycle for the specified number of epochs, we train the model once while actively sampling image pairs every few epochs. This reduces the overhead to train the model multiple times for a large number of epochs. This strategy is followed for every AL approach compared. Specifically, we train all our models (USL and AL) for 50 epochs only, and we propose to actively sample the image pairs after every 10 epochs, resulting in a total of 5 AL cycles. Moreover, all AL baselines perform pseudo-label refinement using NP3 for a fair comparison. Unlike person Re-ID, NIS proposed defining the budget using a fraction of total number of all possible pairwise combinations instead of total number of individuals. We also use a similar technique and use a budget $0.02\%$ in each AL cycle. In all our experiments, we fix the hyperparameters $(\epsilon, k_{max}, s_{min})$ to $(0.6, 5, 0.3)$.

\begin{table*}[!t]\centering
\small
\begin{tabular}{l|cccccccc|c}\toprule
    \textbf{Methods} &\textbf{mAP} &\textbf{mINP} &\textbf{BAKS} &\textbf{AUC ROC} &\textbf{Top-1} &\textbf{Top-3} &\textbf{Top-5} &\textbf{Top-10} &\textbf{Budget} \\  \hline
    ResNet-50 (Baseline) & 38.88\%& 21.34\%& 48.62\%& 60.87\%& 48.37\%& 65.21\%& 71.01\%&81.05\%  & 0\% \\ 
    ResNet-50 (Skyline) & 64.29\%& 49.75\%& 72.04\%& 80.02\%& 72.88\%& 81.06\%& 85.38\%&91.27\%  & 100\% \\ \hline
    \multicolumn{10}{l}{\textit{\textbf{Foundation Methods}}}\\
    MegaDescriptor & 45.65\% &	26.57\% &	55.57\% &	64.12\% &	56.46\% & 70.83\% &	78.14\% &	85.48\%  & 0\%\\
    MiewID & 41.16\%	& 21.08\%	& 57.04\%	& 63.72\%	& 56.41\%	& 69.88\%& 76.63\% &	85.28\%  & 0\% \\ \hline
    \multicolumn{10}{l}{\textit{\textbf{Unsupervised Learning (USL) Methods}}}\\
    ICE & 48.70\%	& 30.08\% &	61.17\% &	70.55\%	& 61.52\%	& 73.93\% & 80.2\% &	87.32\%  & 0\% \\
    PPLR & 51.23\% &	32.62\% & 	61.99\% & 	71.49\% &	63.78\% &	78.19\% & 81.25\% &	87.89\%  & 0\% \\
    ClusterContrast & 45.92\% &	27.81\% & 	57.97\%	& 69.43\% &	58.15\% &	71.83\% & 77.74\% &	86.49\%  & 0\% \\
    RTMem & 48.00\% &	29.28\% &	60.57\% &	69.70\% &  61.4\% &	72.23\% & 77.65\% &	84.85\%  & 0\% \\
    PurificationReID & 51.15\% &	32.63\% &	62.86\% &	72.40\% & 63.16\% & 76.92\% &	83.25\% &	88.45\%   & 0\% \\ \hline
    \multicolumn{10}{l}{\textit{\textbf{Active Learning Methods (based on SpCL)}}}\\
    SpCL (Base)&44.95\% &27.37\% &55.38\% &67.01\% &56.65\% & 71.04\% &76.35\% &84.32\%  & 0\% \\
    + Random &46.67\% &29.02\% &57.28\% &68.33\% &58.11\% & 71.96\% &78.35\% &85.03\%  & 0.1\% \\
    + GBAS &48.61\% &30.35\% &60.53\% &71.50\% &60.56\% & 73.89\% &80.06\% &87.41\%  & 0.1\% \\
    + LBAS &52.15\% &34.06\% &63.62\% &73.15\%  &63.99\% & 76.64\% &82.44\% &88.18\%  & 0.1\% \\
    + NIS &50.72\% &32.73\% &61.21\% &72.30\% &61.78\% & 75.58\% &80.44\% &87.41\%  & 0.1\% \\
    + AAS &\textbf{56.14\%} &\textbf{38.17\%} &\textbf{67.15\%} &\textbf{75.21\%} &\textbf{67.71\%} & \textbf{79.08\%} &\textbf{85.04\%} &\textbf{91.18\%}  & \textbf{0.033\%} \\
    \bottomrule
\end{tabular}
\caption{Averaged results for 13 animal Re-ID datasets. Each block reports performance using foundational, USL and AL Re-ID methods, respectively. SpCL is the base method used for training the Re-ID model after each AL cycle.}\label{tab:animal-results}
\end{table*}

\begin{figure*}[t!]
    \centering
    \begin{subfigure}[t]{0.33\textwidth}
        \centering
        \includegraphics[width=0.86\textwidth]{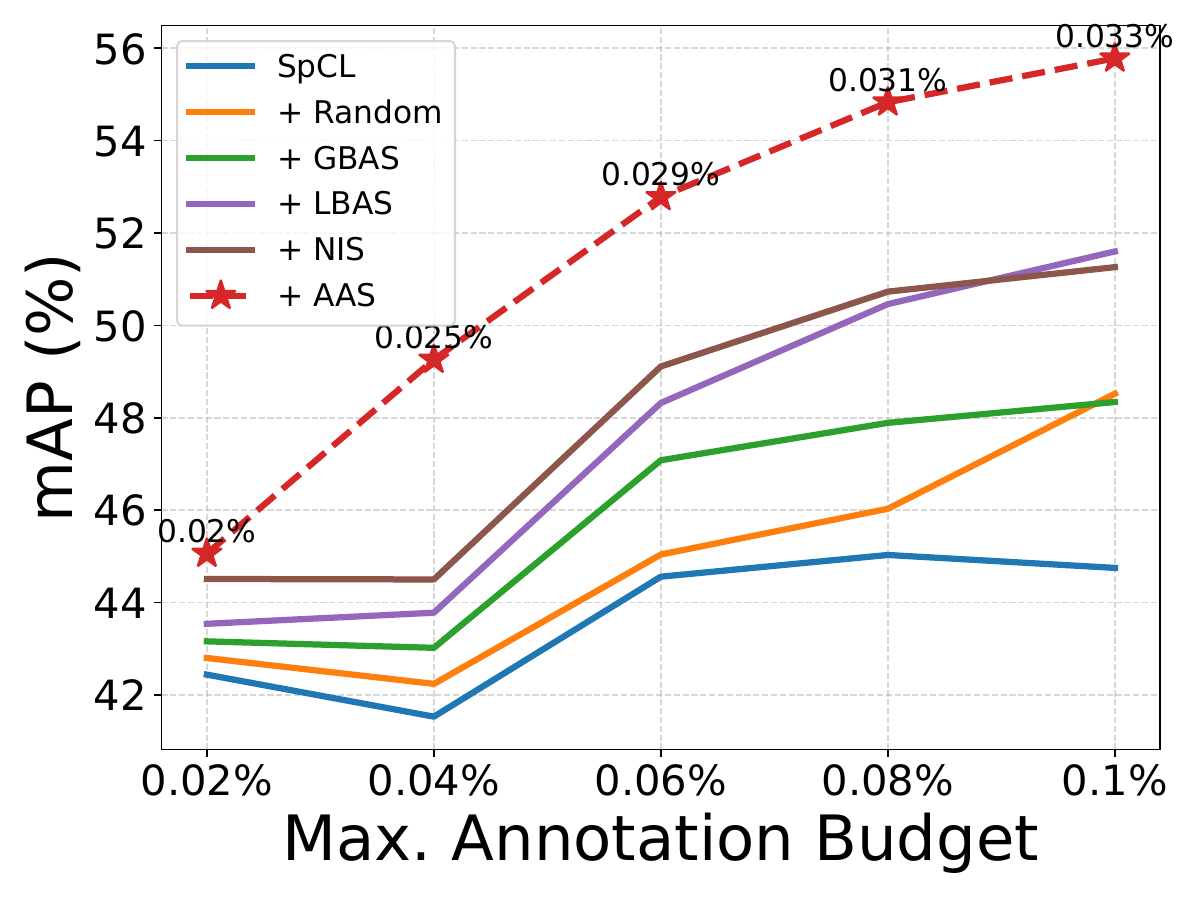}
        \caption{Wildlife Datasets}
    \end{subfigure}
    \begin{subfigure}[t]{0.33\textwidth}
        \centering
        \includegraphics[width=0.86\textwidth]{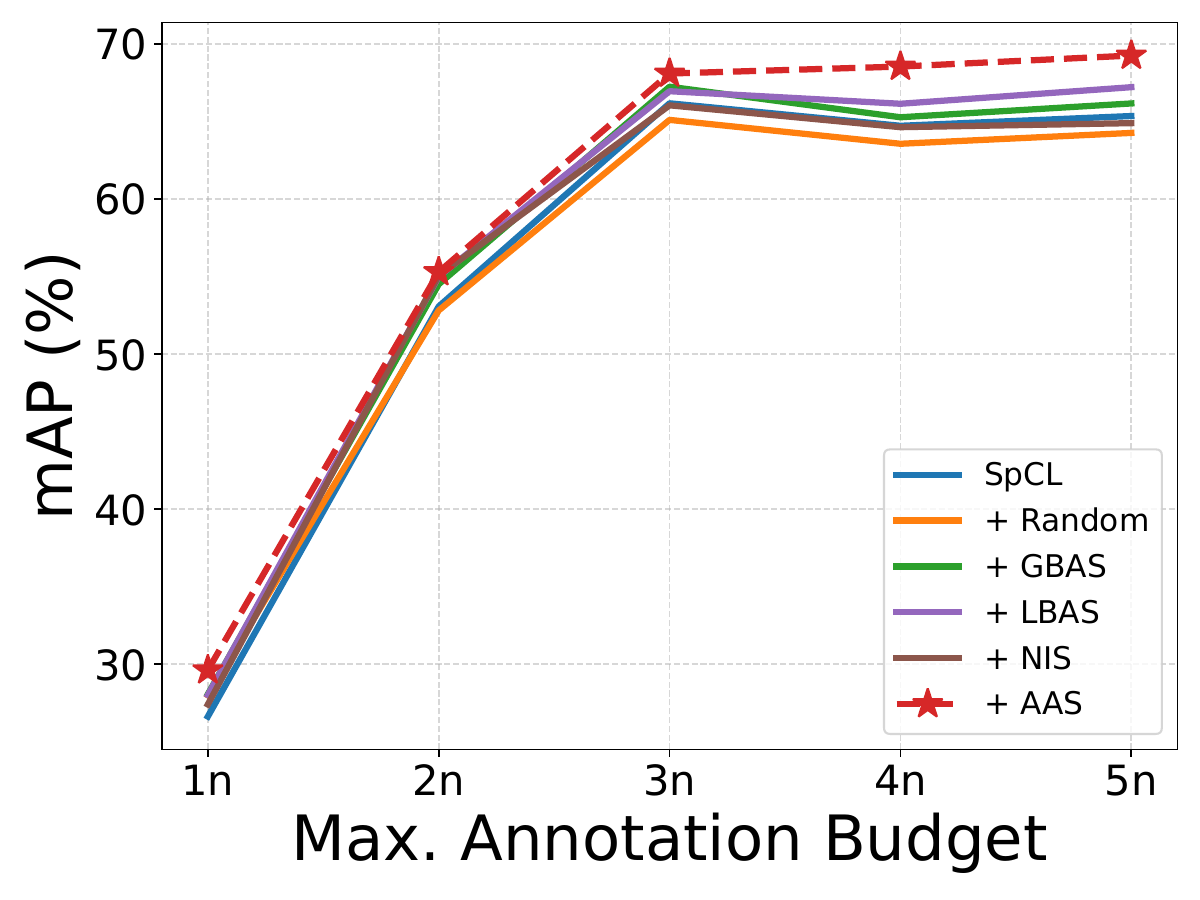}
        \caption{Market-1501}
    \end{subfigure}
    \begin{subfigure}[t]{0.33\textwidth}
        \centering
        \includegraphics[width=0.86\textwidth]{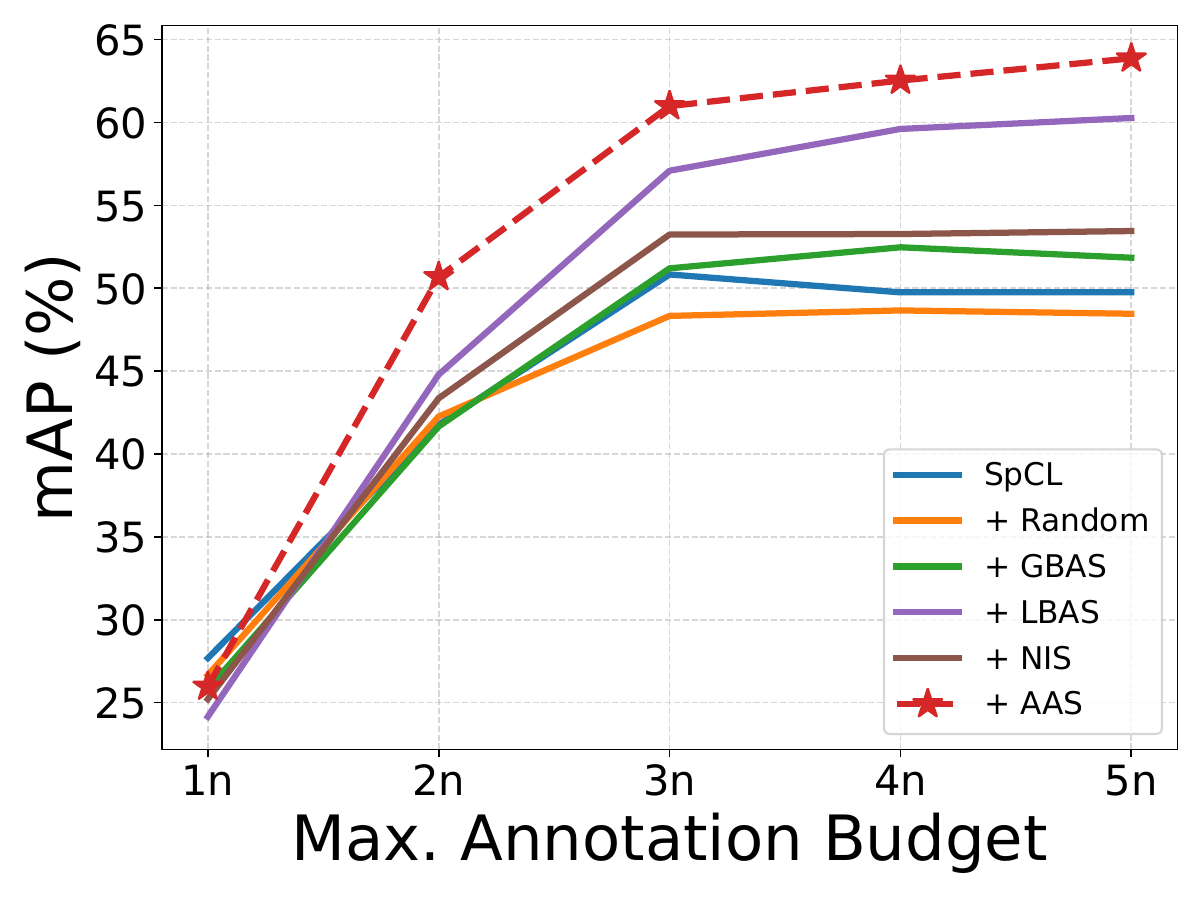}
        \caption{Person-X}
    \end{subfigure}
    \caption{Performance comparison between base SpCL, existing AL Re-ID methods and AAS after every AL cycle. The numbers for AAS in (a) represent the total utilized budget. Since the utilized budget is approximately the same for all AL methods in (b) and (c), we remove them for clarity.}
    \label{fig:al-cycles}
\end{figure*}

\section{Results and Analysis}

The experimental results demonstrate the superior performance of our proposed Ambiguity-Aware Sampling across both wildlife and person Re-ID benchmarks.
We summarize our results on all 13 animal Re-ID datasets in Table \ref{tab:animal-results}, where we report the average performance over 4 runs for all AL Re-ID methods. 
AAS consistently achieves the highest scores across all metrics for closed and open set evaluation, significantly outperforming existing foundational, USL and AL Re-ID methods.
We report that by using only $0.033$\% of all pairwise annotations, we achieve a significant average improvement of $10.49$\%, $11.19$\% and $3.99$\% (mAP) over foundation, USL and AL methods, respectively, while attaining state-of-the-art performance on each dataset. Furthermore, we also show an improvement of $11.09$\%, $8.2$\% and $2.06$\% (AUC ROC) for unknown individuals in an open-world setting.
Notably, while all AL methods were allotted a maximum budget of 0.1\%, AAS inherently utilized only 0.033\% due to its design, further showcasing its efficiency in reducing annotation effort.
This is particularly impressive when compared to the 100\% budget of the Skyline model, as AAS still closely approaches its performance in key metrics like Top-5 accuracy ($85.04\%$ for AAS vs. $85.38\%$ for Skyline), a performance level that other AL methods evidently fail to achieve. In the supplementary, we also show that AAS outperforms LBAS and achieves near skyline performance of $61\%$ mAP with more AL cycles and training iterations, while still sampling only $0.037\%$ of the pairs, even when a much higher budget of $10\%$ is available. We also performed a Wilcoxon signed-rank test on mAP values across all wildlife datasets, obtaining a p-value of at most 1.22e-4 when comparing AAS with all other AL baselines. This highlights that our improvements are statistically significant.

In Fig \ref{fig:al-cycles}, we demonstrate that we also outperform all AL-ReID and base SpCL methods after every AL cycle, showcasing that the pairs sampled by AAS are indeed informative. Furthermore, the budget utilized by AAS in each AL cycle reduces as the training progresses. This indicates that the pseudo-labels in the initial phase of training are noisier and therefore require more supervision. We also provide an analysis of the performance for each of the 13 wildlife datasets in the supplementary, with plots on average performance along with the standard deviation. 

Our analysis on person Re-ID, presented in the supplementary, reinforces the robustness of AAS. We again observe that we consistently outperform all the evaluated methods across both Market-1501 and Person-X datasets. On Market-1501, we observe an improvement of 1.95\% in mAP and 1.05\% in Top-1 accuracy over the best performing and state-of-the-art AL strategy LBAS. Similarly, for the Person-X dataset, we observe an improvement of 3.77\% and 2.68\% in terms of mAP and Top-1 accuracy, respectively, as compared to LBAS. Collectively, these results establish AAS as a highly effective AL strategy that significantly boosts performance while drastically reducing the labeling budget required for both wildlife and person Re-ID applications.

\section{Discussion and Conclusion}
In this work, we presented a novel AL Re-ID framework, AAS, that targets resolving both \textit{over-} and \textit{under-} segmentation errors by leveraging clustering disagreements.
We also introduced NP3, a general-purpose, non-parametric constrained clustering algorithm that seamlessly integrates pairwise constraints into any clustering pipeline.
Our experiments demonstrate that AAS outperforms existing foundation, USL and AL Re-ID techniques on a wide range of wildlife and person Re-ID benchmarks. Importantly, AAS achieves this with only $0.033\%$ budget—highlighting its efficiency and scalability.
Our work highlights the promise of structured uncertainty modeling in AL and contributes a practical, effective framework for real-world Re-ID systems.

\section*{Acknowledgment}
This work was supported by ANRF (SERB), Govt. of India, under grant no. CRG/2020/006049. The authors acknowledge the compute infrastructure support from the Infosys Center for Artificial Intelligence at IIIT-Delhi. The authors would like to thank Anjaneya Sharma from IIIT-Delhi for his support in running additional analysis experiments. Lastly, the authors are grateful to the team at Tiger Cell, Wildlife Institute of India, Dehradun, India, for the collaboration that led to this work. 

\bibliography{bibliography}

\input{Sections/ReproducibilityChecklist}
\end{document}